\documentclass[pdflatex,sn-vancouver-num]{sn-jnl}

\usepackage{graphicx}%
\usepackage{multirow}%
\usepackage{amsmath,amssymb,amsfonts}%
\usepackage{amsthm}%
\usepackage{mathrsfs}%
\usepackage[title]{appendix}%
\usepackage{xcolor}%
\usepackage{textcomp}%
\usepackage{manyfoot}%
\usepackage{booktabs}%
\usepackage{siunitx}%
\usepackage{hyperref}

\theoremstyle{thmstyleone}%

\theoremstyle{thmstyletwo}%

\theoremstyle{thmstylethree}%

\newcommand{\ci}[3]{\shortstack{#1\\[1pt]{\scriptsize(#2 to #3)}}}

\begin{document}

\title[Suicide risk detection in Arabic helpline calls]{Assessing Suicide Risk in Arabic Crisis Helpline Calls: A Comparison of Arabic and English Large Language Models}

\author[1]{\fnm{Linhai} \sur{Ma}}

\author[2]{\fnm{Rita} \sur{El Hachem}}\email{ree21@mail.aub.edu.lb}

\author[3]{\fnm{Mahatab} \sur{El Hajj}}\email{mahatab.elhaj@embracelebanon.org}

\author[2]{\fnm{Lilian} \sur{Ghandour}}\email{lg01@aub.edu.lb}

\author*[1]{\fnm{Samah} \sur{Fodeh}}\email{samah.fodeh@yale.edu}

\affil[1]{\orgdiv{Department of Emergency Medicine},
\orgname{Yale University}, \orgaddress{\city{New Haven}, \state{CT}, \country{USA}}}

 \affil[2]{\orgdiv{Department of Epidemiology and Population Health}, \orgname{American University of Beirut}, \orgaddress{\city{Beirut}, \country{Lebanon}}}

 \affil[3]{\orgdiv{Embrace}, \orgname{Mental Health Center}, \orgaddress{\city{Beirut}, \country{Lebanon}}}

\abstract{\textbf{Background:} Crisis helplines assess suicide risk using structured interviews given by operators. The process takes time and depends on operator training and workload. Natural language processing could support suicide-risk assessment and prioritization. Almost no work has looked at Arabic-language helpline calls, and few studies work within the privacy limits of real helpline data.

\textbf{Methods:} We analysed de-identified transcripts from Lebanon's National Lifeline for Emotional Support and Suicide Prevention. The audio recordings stayed at the helpline. Calls were transcribed on site using a speech recognition model designed for Levantine Arabic. An Arabic named-entity recognition model then removed names and other identifying information, locally. Only the de-identified Arabic transcripts were shared with the research team. Operators recorded the five suicidal ideation items from the Columbia Suicide Severity Rating Scale. We combined these items into two binary outcomes: at-risk and high-risk. We also translated the transcripts into English. This allowed us to compare models on the original Arabic text with models on the English translations. We fine-tuned five instruction-tuned large language models on each corpus, together with six transformer encoders as baselines, four for Arabic and two for English. We evaluated all models on a held-out test set.

\textbf{Results:} We included 383 calls. Of these, 373 were available for the at-risk task, and 52.3\% were positive. A total of 297 calls were available for the high-risk task, and 30.0\% were positive. The best Arabic model reached a macro-F1 of 81.19 and a ROC-AUC of 90.61 for high risk. The best English model reached a macro-F1 of 85.00 and a ROC-AUC of 92.59, and identified 88.9\% of the high-risk calls. The best model in each language separated high-risk calls more cleanly than at-risk calls. Translation to English did not reduce the best observed performance.

\textbf{Conclusion:} The model can classify calls into two suicide risk groups using de-identified Arabic transcripts. This can be done without sending audio outside the helpline. The results for high-risk calls support further testing of the model as a tool for operators. Lower-severity suicidal thoughts were the harder of the two to separate, not the more severe ones. }


\maketitle

\section{Introduction}\label{sec:intro}

About 727\,000 people died by suicide in 2021 \cite{who2025}. Helplines reach people at a useful moment, often while the person is still in crisis and before any clinic has seen them. 
The calls also seem to help in themselves. One study found that callers were less distressed and less suicidal by the end of a call \cite{gould2007}. An evaluation of the Lebanese helpline studied here found the same drop in distress \cite{zeinoun2021}.
However, judging risk on a helpline is still hard. Operators work through a structured questionnaire, usually the Columbia Suicide Severity Rating Scale (C-SSRS) \cite{posner2011}. They do this while the call is going on, and under time pressure. A systematic review of such questionnaires found that none of them is accurate enough to be used on its own \cite{sbu2015}. The same review found no good evidence yet on whether adding one to the clinician's own judgement helps. Staff turnover makes the problem worse, because good assessment depends on training that is expensive to keep up.

Natural language processing (NLP) could help here, and a growing body of work applies it to suicide risk \cite{arowosegbe2023,castillo2020,fodeh2017}. Two gaps stand out in that work. Most of it uses social media posts or clinical notes, so the text is not the kind of text a real tool would see. Almost all of it is also in English, and reviews have asked for work in other languages and inside real services \cite{castillo2020}. 
Earlier helpline studies cover services that answer in English \cite{broadbent2023,iyer2022} and in Chinese \cite{su2025,wang2025}. One service answers in both Hebrew and Arabic, but its published analysis used only the Hebrew chats \cite{grimland2025}.
Arabic is one clear gap. Its spoken dialects differ from Modern Standard Arabic in exactly the informal way people talk on a crisis call. How people speak about suicide also depends on religious and social taboos, which vary from place to place.

Lebanon has both the need and the data. National surveys report high rates of mental disorder among young people, together with very little help-seeking \cite{maalouf2022,baroud2019}. The first nationwide study of suicide deaths there found wide variation between regions and groups, set against strong social taboo \cite{bizri2021}. Embrace runs the National Lifeline for Emotional Support and Suicide Prevention with the Ministry of Public Health's National Mental Health Programme. It is the only suicide prevention helpline in the country. Its calls are recorded, and its operators complete the C-SSRS under supervision. The service therefore holds recorded speech with matching assessments, which is what model development needs. This is also very sensitive data. Callers are in crisis, they mention identifying details, and they have been promised confidentiality. A recording of someone's voice identifies them, and unlike text it cannot be edited to hide that. Any analysis therefore has to keep the audio inside the service. This limit is usually treated as an obstacle to research. We treated it as a requirement and designed the pipeline around it. We set out to build and test a model that sorts calls into risk categories from de-identified Arabic transcripts alone. We also wanted to know what it costs to translate the transcripts into English first, since translate then classify is the usual approach for a language with few resources. Beyond that, we wanted to describe how these models fail when the risk labels are imbalanced, because that is what decides whether a model is safe to put in front of an operator.

This study makes three contributions. First, it builds and tests a risk classifier for Arabic-language helpline calls. 
Second, it describes a working pipeline where speech recognition and de-identification run inside the service, so the recordings never move. Third, it compares Arabic text with an English translation of the same calls, using the same splits and the same labels. 

\subsection{Relatedwork}

Most NLP work on suicide risk uses social media posts or clinical notes \cite{arowosegbe2023,castillo2020,fodeh2017}. Helpline calls are different in ways that matter. They are conversations rather than written statements, and the risk assessment happens during the call instead of being pieced together afterwards. The label also comes from a trained operator at the time, not from later coding. Several groups have worked on helpline contacts directly. One study classified risk levels in text-based crisis counselling. It found that a neural model missed fewer at-risk clients than a term-frequency model \cite{broadbent2023}. Two other studies classified low against high risk from call audio rather than from words. One used voice biomarkers on an Australian telehealth service, with counsellor ratings re-checked against the C-SSRS \cite{iyer2022}. The other used acoustic features on a Chinese hotline \cite{su2025}. A further study on a Chinese hotline combined pitch features with deep learning features to analyse the emotion expressed during calls \cite{wang2025}. 
The Sahar helpline answers in both Hebrew and Arabic, but its published analysis used only the Hebrew chats \cite{grimland2025}. That gap shows how little is known about Arabic crisis calls. The helpline studied here has been analysed before. One study evaluated whether calls reduced caller distress \cite{zeinoun2021}, and another modelled which caller characteristics went with suicidal ideation, intent and self-harm \cite{farran2025}. Both used the structured fields the operator fills in, not the speech. As far as we can tell, two things here are new.
1) This is the first suicide-risk classifier we know of for Levantine Arabic-language crisis helpline calls. 2) It is also the first report of a full pipeline where speech recognition and de-identification run inside the service, so the recordings are never moved. Reviews of this literature have asked for work outside English and inside real services \cite{castillo2020}. Our results agree with the general finding that transformer models pick up suicidal content well above chance. But we also show that performance varied substantially across both outcome definitions and model backbones.

\section{Methods}\label{sec:methods}

\subsection{Settings, data source and ethical approval}\label{subsec:setting}

Data came from calls to the Embrace National Lifeline (1564) in Lebanon. Trained operators answer the calls, supervised by clinical psychologists. They complete a standard C-SSRS assessment and record the result in a service database. The helpline is anonymous, so call records cannot be linked to national identifiers. This was a secondary analysis of routine service data. We had no contact with callers. Ethical approval was in place before the data were accessed. A call was eligible if the recording lasted between five and ten minutes. This window drops very short contacts, which are mostly wrong numbers, silent calls and immediate hang-ups. Section~\ref{subsec:limitations} discusses what this leaves out.

\subsection{Privacy-preserving transcription and de-identification}\label{subsec:pipeline}

Recordings did not leave the helpline. All speech processing ran on site, and only redacted Arabic text went to the analysis team. Recordings were transcribed with a Whisper-family speech recognition model \cite{radford2023}. A call is longer than the model's input window, so we split each recording into overlapping segments and removed the duplicated text when joining the segments back together. The calls are in Levantine Arabic rather than Modern Standard Arabic, so we used a decoding setup adapted to the dialect. The transcripts then went through an Arabic named-entity recognition model, which removed names of people, places and organisations along with other direct identifiers. Phone number blocks were stripped from the filenames before release. The whole code for the transcription and de-identification is available online (see the Section Code availability).

\subsection{Outcome definition}\label{subsec:outcome}

Operators record the five ideation items of the C-SSRS ladder: wish to be dead; non-specific active suicidal thoughts; active ideation with any method; active ideation with some intent to act; and active ideation with a specific plan and intent \cite{posner2011}. Each item is marked present or absent. Other codes cover items that were not asked, items skipped by the logic of the questionnaire, and items the caller did not answer. We grouped the items into two categories by severity, following the order of the ladder. The two lower rungs, a wish to be dead and non-specific active thoughts, make up the \emph{at-risk} category. The three items that involve a method, an intent or a plan make up \emph{high risk}:
\begin{align}
y^{\mathrm{at}} &= y^{\mathrm{WD}} \vee y^{\mathrm{NA}}, \label{eq:atrisk}\\
y^{\mathrm{high}} &= y^{\mathrm{AM}} \vee y^{\mathrm{IA}} \vee y^{\mathrm{PI}}. \label{eq:high}
\end{align}
A call is at-risk if either of the two lower items is present, and high risk if any of the three higher items is present. The two categories cover overlapping but different sets of calls, and we modelled them as two separate yes-or-no tasks. 

\subsection{Handling missing annotation}\label{subsec:missing}

Not every call has all five items recorded. A missing item does not always leave the category unclear. If any recorded item is present, the category is positive whatever the missing items would have said. The category is unclear only when every recorded item is absent and at least one item was never recorded, because then the missing item alone decides the answer. We kept the calls whose category was clear under this rule and dropped the rest. This resulted in the exclusion of \num{32} calls from the at-risk category and \num{16} from the high-risk category. Treating unrecorded items as absent would have allowed these calls to remain in the analysis, but would risk introducing false negatives into a corpus designed to detect suicide risk. This could bias the models toward missing positive cases, which is the more consequential error in this setting.

\subsection{Machine translation}\label{subsec:mt}

To compare modelling the source language against translate then classify, we translated each de-identified Arabic transcript into English with Qwen/Qwen2.5-72B-Instruct model under greedy decoding. We then checked that each English output covered the whole call. \num{29} of the \num{438} transcripts were ill-formed or incomplete, so we dropped them. We removed these calls from both corpora rather than from the English one alone. Dropping them only on the English side would leave the two languages covering different sets of calls, which would have affected the comparison between languages.

\subsection{Model Training}\label{subsec:models}

Table~\ref{tab:models} lists every model we fine-tuned. We selected five instruction-tuned large language model (LLM) backbone \textbf{decoders} covering a range of size and pretraining language, and six transformer \textbf{encoders} (BERT-based Models) for comparison. To fine-tune these models, we turned each call into an instruction-tuning triple. The instruction states the yes or no question, the input is the full transcript, and the target output is a single word. Every model used the same 4-bit QLoRA recipe with no per-model hyperparameter search, so that differences between models reflect the backbone rather than tuning effort \cite{hu2022,dettmers2023}. We attached low-rank adapters ($r=16$, $\alpha=32$, dropout 0.05) to the attention and feed-forward projections. Base weights were frozen and quantised. Training minimised causal language-model cross-entropy on the answer token only, with prompt tokens masked, using a cosine schedule (peak learning rate $2\times10^{-4}$, effective batch size 16). We trained separate adapters per model, task and language, and ran both a three-epoch and a ten-epoch schedule. At inference we decoded the answer token greedily and mapped it to a binary label. Generations that could not be parsed were counted as negative. Zero-shot baselines used the same prompt on the 0-Shot base model. Every transcript is several times longer than a 512-token encoder window, so we read documents in overlapping windows and pooled the window representations, following standard practice for long documents \cite{pappagari2019}.

\begin{table}[!ht]
\caption{Models fine-tuned in this study. Each model was fine-tuned/tested separately for both tasks on the corpus listed in the final column. Decoders answer the risk question with a single word. Encoders read each transcript in overlapping windows. Parameter counts are the published configurations.}\label{tab:models}
\footnotesize
\setlength{\tabcolsep}{4pt}
\begin{tabular}{@{}lllll@{}}
\toprule
Model & Class & Parameters & Pretraining emphasis & Corpus \\
\midrule
Qwen2.5-1.5B-Instruct \cite{qwen2025}   & Decoder & 1.5B & General, multilingual & Arabic, English \\
Qwen2.5-14B-Instruct \cite{qwen2025}    & Decoder & 14B  & General, multilingual & Arabic, English \\
Llama-3.3-70B-Instruct \cite{llama2024} & Decoder & 70B  & General, multilingual & Arabic, English \\
AceGPT-v2-8B-Chat \cite{acegpt2024,zhu2025}  & Decoder & 8B  & Arabic-centric & Arabic, English \\
AceGPT-v2-70B-Chat \cite{acegpt2024,zhu2025} & Decoder & 70B & Arabic-centric & Arabic, English \\
\midrule
CAMeLBERT-DA \cite{inoue2021}      & Encoder & 108M & Arabic, dialectal   & Arabic \\
AraBERTv0.2 \cite{antoun2020}      & Encoder & 135M & Arabic              & Arabic \\
MARBERT-based \cite{abdulmageed2021} & Encoder & 163M & Arabic, dialectal & Arabic \\
Multilingual BERT \cite{devlin2019} & Encoder & 178M & Multilingual       & Arabic \\
BERT-base \cite{devlin2019}        & Encoder & 110M & English            & English \\
BERT-large \cite{devlin2019}       & Encoder & 340M & English            & English \\
\botrule
\end{tabular}
\end{table}

\begin{table}[!t]
\caption{Derivation of the analysis sample. The two categories draw on different sets of calls, because the five ideation items were not recorded on a common set. Each category starts from the calls with at least one of its own constituent items recorded. Calls were then dropped because the category could not be determined from the recorded items, or because the machine translation was incomplete. The two categories overlap, so a call with an incomplete translation can be counted in both branches; \num{29} distinct calls were affected.}\label{tab:flow}
\begin{tabular}{@{}lr@{}}
\toprule
Stage & Calls \\
\midrule
De-identified transcripts produced & 987 \\
Matched to a service record & 545 \\
\quad Matched to more than one record & 171 \\
\quad\quad Kept: records agreed on all items & 154 \\
\quad\quad Dropped: records disagreed & 17 \\
Transcripts with unambiguous labels & 528 \\
With $\geq$1 recorded ideation item & 438 \\
\quad of which with all five items & 280 \\
\quad of which with an incomplete translation & 29 \\
\midrule
\multicolumn{2}{@{}l}{\textit{At-risk}} \\
\quad With $\geq$1 of its two items recorded & 433 \\
\quad Dropped: category undetermined & 32 \\
\quad Dropped: incomplete translation & 28 \\
\quad Analysable & 373 \\
\quad\quad Training / test & 298 / 75 \\
\midrule
\multicolumn{2}{@{}l}{\textit{High risk}} \\
\quad With $\geq$1 of its three items recorded & 333 \\
\quad Dropped: category undetermined & 16 \\
\quad Dropped: incomplete translation & 20 \\
\quad Analysable & 297 \\
\quad\quad Training / test & 237 / 60 \\
\midrule
Distinct calls in at least one category & 383 \\
\botrule
\end{tabular}
\end{table}

\begin{table}[t]
\caption{Transcript length in whitespace-delimited words. The two corpora hold the same calls, so the Arabic and English rows are paired.}\label{tab:length}
\begin{tabular}{@{}llrrrrrr@{}}
\toprule
Language & Split & $n$ & Median & Mean & P25 & P75 & Max \\
\midrule
\multirow{5}{*}{Arabic}
 & At-risk, training & 298 & 730 & 756 & 597 & 881 & 1381 \\
 & At-risk, test     & 75  & 731 & 749 & 608 & 885 & 1170 \\
 & High, training     & 237 & 731 & 761 & 600 & 892 & 1641 \\
 & High, test         & 60  & 692 & 702 & 577 & 829 & 1146 \\
 & All calls          & 383 & 732 & 759 & 602 & 885 & 1641 \\
\midrule
\multirow{5}{*}{English}
 & At-risk, training & 298 & 928 & 953 & 762 & 1085 & 1655 \\
 & At-risk, test     & 75  & 944 & 945 & 764 & 1097 & 1449 \\
 & High, training     & 237 & 935 & 959 & 764 & 1096 & 2269 \\
 & High, test         & 60  & 888 & 888 & 715 & 1069 & 1392 \\
 & All calls          & 383 & 934 & 957 & 763 & 1097 & 2269 \\
\botrule
\end{tabular}
\end{table}

\subsection{Evaluation}\label{subsec:analysis}

We split at the call level, stratified on the category label, with an 80:20 training to test ratio. The split was defined over transcript identifiers and applied the same way to the Arabic corpus and to its English translation. The two datasets therefore hold the same calls, in the same splits, with the same labels, and differ only in language. No call appears on both sides of the split. For evaluation, we used the area under the receiver operating characteristic curve (ROC-AUC), the area under the precision and recall curve (PR-AUC), macro-averaged F1, positive-class recall and overall accuracy. Positive-class recall is the same thing as sensitivity to the positive class. Macro-F1 here is the mean of the two per-class F1 scores, so it need not fall between macro-averaged precision and recall. All values are percentages.  Table~\ref{tab:operating} reports sensitivity, specificity and the two predictive values, together with the two-by-two table of correct and incorrect decisions they derive from. Each decoder answers yes or no, so the threshold is whatever the model itself applies.
ROC-AUC and PR-AUC need a continuous score rather than a decision, so for those two measures we used the probability of the positive answer token, normalised over the yes and no tokens. Proportions in Table~\ref{tab:operating} are given with a 95\% Wilson score interval, which behaves better than a normal approximation when the counts are small. We report accuracy for completeness, but we did not use it to pick or compare models. With 30\% to 52\% of calls positive, accuracy is partly inflated by guessing the majority class, which is the behaviour this analysis is meant to catch. The study used one stratified split. With \num{75} and \num{60} test calls, moving one call changes macro-F1 by about one to two points. We report the rankings descriptively and draw no statistical comparison between models. The intervals in Table~\ref{tab:operating} show how wide the uncertainty is. We report ROC-AUC and PR-AUC without intervals and treat them as descriptive.

\section{Results}\label{sec:results}

Of \num{987} de-identified transcripts, \num{545} matched a service record by call identifier. Among these, \num{171} identifiers matched more than one service record. Of those, \num{154} carried identical ideation labels across records and were kept, while \num{17} carried conflicting labels and were dropped, leaving \num{528} transcripts with unambiguous labels. Of these, \num{438} had at least one ideation item recorded, and \num{280} had all five. The five items were not recorded on a common set of calls, so each category was derived from its own base: \num{433} calls had at least one of the two at-risk items recorded, and \num{333} had at least one of the three high-risk items. Applying the rule in Section~\ref{subsec:missing} and removing calls with an incomplete machine translation left \num{373} analysable calls for the at-risk category and \num{297} for high risk, drawn from \num{383} distinct calls in total. Table~\ref{tab:flow} shows the derivation. The at-risk category was close to balanced, at \num{52.3}\% positive in training and \num{52.0}\% in test. High risk was \num{30.0}\% positive in both. Stratification held training and test rates within \num{0.3} percentage points on both tasks. Calls were long (Table~\ref{tab:length}). The median Arabic transcript ran to \num{732} words and the longest to \num{1641}. The English version was longer throughout, with a median of \num{934} words, about \num{28}\% more. Arabic morphology packs into single words what English spreads over several. Any length-based processing budget must therefore be set from the English side.

\subsection{Overall model performance}\label{subsec:overview}

\begin{figure}[!ht]
\centering
\includegraphics[width=\textwidth]{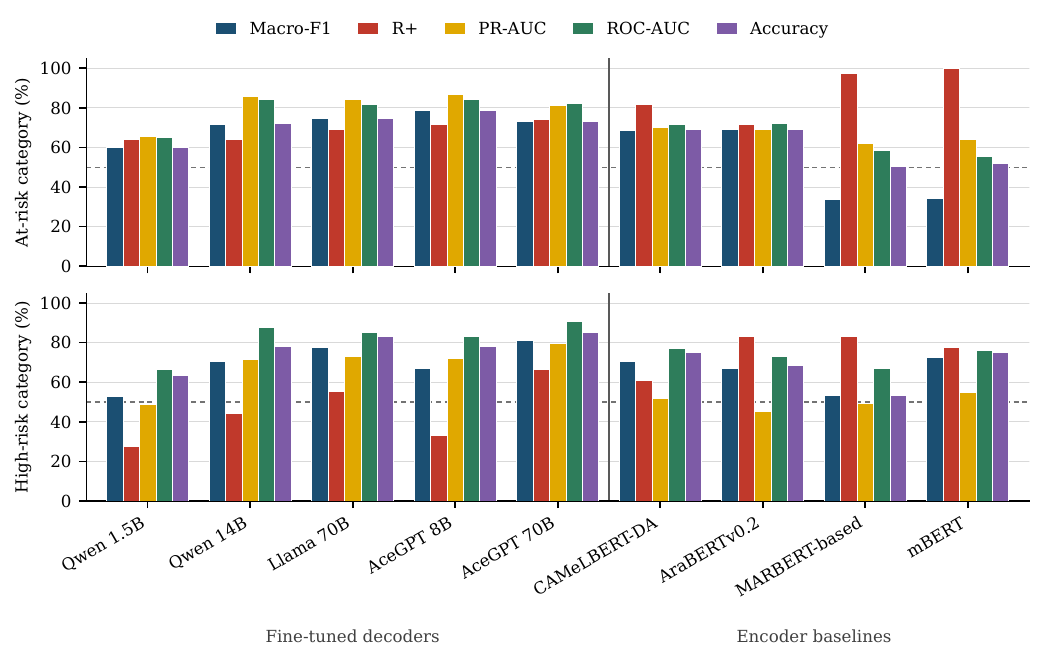}
\caption{All fine-tuned models on the Arabic corpus at ten epochs, on the held-out test split. The upper panel is the at-risk category and the lower panel is the high-risk category. R+ is positive-class recall. The dashed line marks 50 per cent. The five models left of the vertical rule are the decoders and the four to its right are the encoder baselines. Tables~\ref{tab:exploratory} and~\ref{tab:encoders} give the macro-F1, positive-class recall and ROC-AUC values numerically.}\label{fig:arabic}
\end{figure}

\begin{figure}[!ht]
\centering
\includegraphics[width=\textwidth]{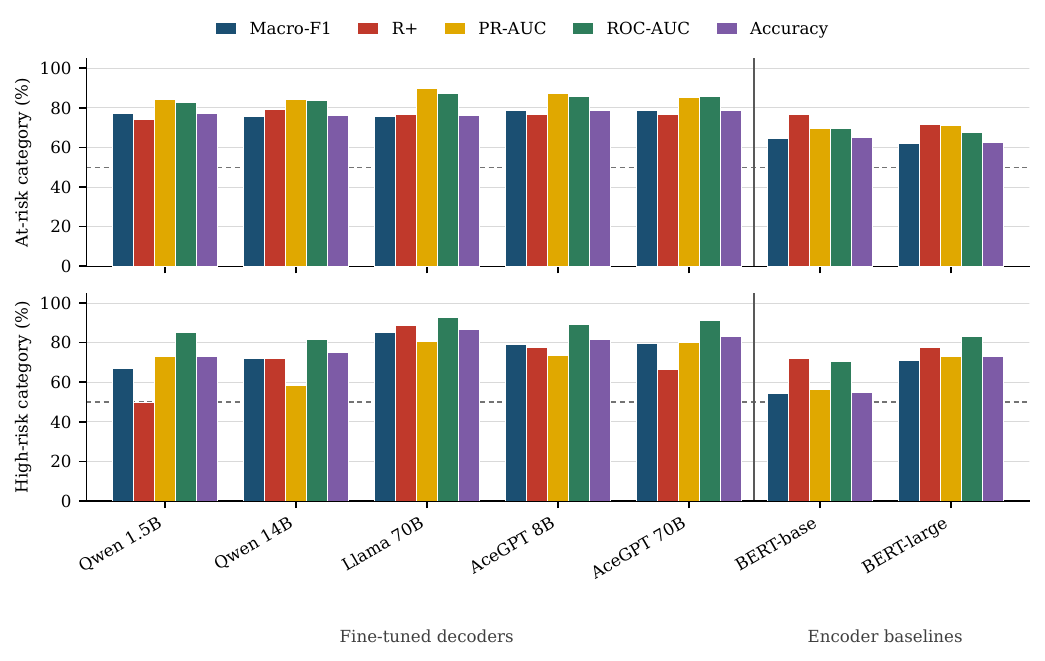}
\caption{All fine-tuned models on the English corpus at ten epochs, on the held-out test split. Panels, measures and layout are as in Figure~\ref{fig:arabic}. The English corpus is the machine translation of the same calls, in the same splits, with the same labels.}\label{fig:english}
\end{figure}

Figures~\ref{fig:arabic} and~\ref{fig:english} show every fine-tuned model on all five measures, one figure per corpus, at ten epochs. Both figures use the same scale, so the two risk categories can be read against each other. Most models sit between 65 and 90 on macro-F1, ROC-AUC and accuracy, while positive-class recall spreads much more widely, from 27.78 to 100.00 on the Arabic corpus. The best encoder falls below the best decoder in both languages and on both categories, and two Arabic encoders sit near the floor on the at-risk category. The spread across decoders is narrower on the English corpus than on the Arabic one, so the choice of backbone matters less once the text has been translated. The rest of this section reads out of these two figures.

\subsection{Primary models' performance}\label{subsec:primary}

Table~\ref{tab:primary} reports one 70B decoder per corpus against its 0-Shot baseline, the Arabic-centric model on Arabic and the general-purpose model on English. For high risk, the Arabic model reached a macro-F1 of \num{81.19} and a ROC-AUC of \num{90.61}, against \num{41.18} and \num{58.93} 0-Shot. The 0-Shot model found none of the high-risk calls in the test split. After fine-tuning it found \num{66.7}\% of them. The English model reached a macro-F1 of \num{85.00} and a ROC-AUC of \num{92.59} and found \num{88.9}\% of high-risk calls, against \num{73.78} and \num{85.45} 0-Shot. For the at-risk category, the Arabic model reached a macro-F1 of \num{73.29} and a ROC-AUC of \num{82.19}. The English model reached \num{75.96} and \num{87.25}. Both improved on their 0-Shot baselines by a wide margin. Those baselines found \num{12.8}\% and \num{38.5}\% of at-risk calls.

\begin{table}[!ht]
\caption{Primary 70B decoders, before and after fine-tuning, on the held-out test split. All values are percentages. R+ is positive-class recall. Table~\ref{tab:operating} gives the operating characteristics for the same eight conditions.}\label{tab:primary}
\begin{tabular}{@{}llrrrrr@{}}
\toprule
Model & category (\% pos) & Macro-F1 & R+ & PR-AUC & ROC-AUC & Accuracy \\
\midrule
\multicolumn{7}{@{}l}{\textit{Arabic corpus, Arabic-centric 70B decoder (ACEGPT 70B)}} \\
\quad 0-Shot    & At-risk (52) & 45.33 & 12.82 & 58.18 & 56.91 & 54.67 \\
\quad Fine-tuned & At-risk (52) & 73.29 & 74.36 & 81.13 & 82.19 & 73.33 \\
\quad 0-Shot    & High (30)     & 41.18 & 0.00  & 38.36 & 58.93 & 70.00 \\
\quad Fine-tuned & High (30)     & 81.19 & 66.67 & 79.55 & 90.61 & 85.00 \\
\midrule
\multicolumn{7}{@{}l}{\textit{English corpus, general-purpose 70B decoder (LLAMA 70B)}} \\
\quad 0-Shot    & At-risk (52) & 56.01 & 38.46 & 68.54 & 64.89 & 57.33 \\
\quad Fine-tuned & At-risk (52) & 75.96 & 76.92 & 89.85 & 87.25 & 76.00 \\
\quad 0-Shot    & High (30)     & 73.78 & 61.11 & 78.75 & 85.45 & 78.33 \\
\quad Fine-tuned & High (30)     & 85.00 & 88.89 & 80.79 & 92.59 & 86.67 \\
\botrule
\end{tabular}
\end{table}

\subsection{Model performance across risk categories}\label{subsec:operating}

Table~\ref{tab:operating} gives sensitivity, specificity and the predictive values for the same eight conditions. PPV is positive predictive value and NPV is the negative predictive value. The two fine-tuned models behaved differently on the high-risk category. The English model found 16 of the 18 high-risk calls, giving a sensitivity of \num{88.9}\% and a negative predictive value of \num{94.7}\%. It also raised 6 false alarms out of 42 negative calls. The Arabic model missed 6 of the 18, so its sensitivity was \num{66.7}\%, but it raised only 3 false alarms. On the at-risk category the two models were close, and both sat near \num{75}\% on all four measures. The intervals are wide throughout. On the high-risk category they are wide enough that the two languages cannot be told apart. With only 18 positive calls in that test split, one call moves sensitivity by more than five percentage points.

\begin{table}[!ht]
\caption{Operating characteristics on the held-out test split, with 95\% Wilson score intervals. Counts are true positives, false positives, false negatives and true negatives. Arabic denotes the Arabic-centric 70B decoder (ACEGPT 70B) on the Arabic corpus. English is the general-purpose 70B decoder (LLAMA 70B) on the English corpus. The conditions are the same eight reported in Table~\ref{tab:primary}. Predictive value is undefined where a model predicted no positive case.}\label{tab:operating}
\footnotesize
\setlength{\tabcolsep}{3pt}
\begin{tabular}{@{}lccccc@{}}
\toprule
Model & TP/FP/FN/TN & Sensitivity & Specificity & PPV & NPV \\
\midrule
\multicolumn{6}{@{}l}{\textit{At-risk category (52\% positive)}} \\
\quad Arabic, 0-Shot    & 5/0/34/36   & \ci{12.8}{5.6}{26.7}  & \ci{100.0}{90.4}{100.0} & \ci{100.0}{56.6}{100.0} & \ci{51.4}{40.0}{62.8} \\
\quad Arabic, fine-tuned & 29/10/10/26 & \ci{74.4}{58.9}{85.4} & \ci{72.2}{56.0}{84.2}   & \ci{74.4}{58.9}{85.4}   & \ci{72.2}{56.0}{84.2} \\
\quad English, 0-Shot   & 15/8/24/28  & \ci{38.5}{24.9}{54.1} & \ci{77.8}{61.9}{88.3}   & \ci{65.2}{44.9}{81.2}   & \ci{53.8}{40.5}{66.7} \\
\quad English, fine-tuned& 30/9/9/27   & \ci{76.9}{61.7}{87.4} & \ci{75.0}{58.9}{86.2}   & \ci{76.9}{61.7}{87.4}   & \ci{75.0}{58.9}{86.2} \\
\midrule
\multicolumn{6}{@{}l}{\textit{High-risk category (30\% positive)}} \\
\quad Arabic, 0-Shot    & 0/0/18/42   & \ci{0.0}{0.0}{17.6}   & \ci{100.0}{91.6}{100.0} & undefined               & \ci{70.0}{57.5}{80.1} \\
\quad Arabic, fine-tuned & 12/3/6/39   & \ci{66.7}{43.7}{83.7} & \ci{92.9}{81.0}{97.5}   & \ci{80.0}{54.8}{93.0}   & \ci{86.7}{73.8}{93.7} \\
\quad English, 0-Shot   & 11/6/7/36   & \ci{61.1}{38.6}{79.7} & \ci{85.7}{72.2}{93.3}   & \ci{64.7}{41.3}{82.7}   & \ci{83.7}{70.0}{91.9} \\
\quad English, fine-tuned& 16/6/2/36   & \ci{88.9}{67.2}{96.9} & \ci{85.7}{72.2}{93.3}   & \ci{72.7}{51.8}{86.8}   & \ci{94.7}{82.7}{98.5} \\
\botrule
\end{tabular}
\end{table}

\subsection{Better separation for more severe cases}\label{subsec:severity}

The best model in each language scored higher on the high-risk category than on the at-risk category, even though high risk is the rarer outcome. The best Arabic figures were \num{81.19} macro-F1 for high risk against \num{78.63} for the at-risk category. The best English figures were \num{85.00} against \num{78.66} (Table~\ref{tab:exploratory}). ROC-AUC followed the same order. This is the opposite of what class prevalence alone would predict. Smaller backbones did not follow it: on the Arabic corpus the 1.5B, 14B and 8B decoders all scored higher on the at-risk category. Positive-class recall did not follow. Every Arabic decoder found a smaller share of high-risk calls than of at-risk calls, and so did three of the five English ones. On the Arabic corpus the gain on high risk is therefore in separation rather than in how many positive calls are caught at the model's own threshold. The separation itself suggests the main limit is not how rare the label is but how the construct is expressed. Talk of a method, an intent or a plan is direct and easy to quote. Lower-severity ideation is indirect, and its wording overlaps heavily with the wording of general distress, which the helpline also receives in volume.

Table~\ref{tab:exploratory} reports all ten decoder conditions at ten epochs.
Parameter count did not set the order: the 8B Arabic-centric model reached \num{78.63} macro-F1 on the Arabic at-risk category against \num{73.29} for its 70B counterpart. 



\begin{table}[!ht]
\caption{Exploratory comparison of all fine-tuned decoder conditions at ten epochs. R+ is positive-class recall, that is, sensitivity to the risk-positive class. Given the size of the test split these comparisons are descriptive and do not fix an order among backbones.}\label{tab:exploratory}
\footnotesize
\setlength{\tabcolsep}{4pt}
\begin{tabular}{@{}llrrrrrr@{}}
\toprule
Corpus & Backbone & \multicolumn{2}{c}{Macro-F1} & \multicolumn{2}{c}{R+} & \multicolumn{2}{c}{ROC-AUC} \\
\cmidrule(lr){3-4}\cmidrule(lr){5-6}\cmidrule(lr){7-8}
 & & At-risk & High & At-risk & High & At-risk & High \\
\midrule
\multirow{5}{*}{Arabic}
 & 1.5B general      & 59.82 & 53.12 & 64.10 & 27.78 & 65.17 & 66.40 \\
 & 14B general       & 71.92 & 70.44 & 64.10 & 44.44 & 84.37 & 87.70 \\
 & 70B general       & 74.65 & 77.78 & 69.23 & 55.56 & 82.02 & 85.05 \\
 & 8B Arabic-centric & 78.63 & 67.16 & 71.79 & 33.33 & 84.47 & 83.07 \\
 & 70B Arabic-centric& 73.29 & 81.19 & 74.36 & 66.67 & 82.19 & 90.61 \\
\midrule
\multirow{5}{*}{English}
 & 1.5B general      & 77.33 & 67.17 & 74.36 & 50.00 & 83.01 & 85.19 \\
 & 14B general       & 75.89 & 72.21 & 79.49 & 72.22 & 84.01 & 81.88 \\
 & 70B general       & 75.96 & 85.00 & 76.92 & 88.89 & 87.25 & 92.59 \\
 & 8B Arabic-centric & 78.66 & 79.11 & 76.92 & 77.78 & 85.90 & 89.35 \\
 & 70B Arabic-centric& 78.66 & 79.48 & 76.92 & 66.67 & 85.83 & 91.34 \\
\botrule
\end{tabular}
\end{table}

\subsection{BERT Encoder baselines}\label{subsec:encoders}

The best fine-tuned decoder (LLM) beat the best encoder baseline (BERT) on every task and language (Table~\ref{tab:encoders}). The best Arabic encoder score was \num{69.25} macro-F1 on the at-risk category and \num{72.82} on high risk, against \num{78.63} and \num{81.19} for decoders. The best English encoder score was \num{64.57} and \num{71.29}, against \num{78.66} and \num{85.00}. No single encoder was best on both categories in either language.


\begin{table}[!ht]
\caption{Transformer encoder baselines, fine-tuned for ten epochs. R+ is positive-class recall, that is, sensitivity to the risk-positive class. Encoders read each transcript in overlapping windows, since all transcripts exceed the 512-token input limit.}\label{tab:encoders}
\footnotesize
\setlength{\tabcolsep}{4pt}
\begin{tabular}{@{}llrrrrrr@{}}
\toprule
Corpus & Encoder & \multicolumn{2}{c}{Macro-F1} & \multicolumn{2}{c}{R+} & \multicolumn{2}{c}{ROC-AUC} \\
\cmidrule(lr){3-4}\cmidrule(lr){5-6}\cmidrule(lr){7-8}
 & & At-risk & High & At-risk & High & At-risk & High \\
\midrule
\multirow{4}{*}{Arabic}
 & CAMeLBERT-DA      & 68.53 & 70.69 & 82.05 & 61.11 & 71.65 & 77.12 \\
 & AraBERTv0.2       & 69.25 & 67.23 & 71.79 & 83.33 & 72.44 & 72.88 \\
 & MARBERT-based     & 33.63 & 53.28 & 97.44 & 83.33 & 58.40 & 67.06 \\
 & Multilingual BERT & 34.21 & 72.82 & 100.00 & 77.78 & 55.34 & 76.06 \\
\midrule
\multirow{2}{*}{English}
 & BERT-base  & 64.57 & 54.38 & 76.92 & 72.22 & 69.66 & 70.63 \\
 & BERT-large & 62.12 & 71.29 & 71.79 & 77.78 & 67.66 & 83.07 \\
\botrule
\end{tabular}
\end{table}

\subsection{Model performance under class imbalance}\label{subsec:collapse}

Several conditions produced trivial classifiers, and they failed in opposite directions (Table~\ref{tab:collapse}). The smallest decoder on Arabic high risk, trained for three epochs, answered negative to every call, giving \num{0.00} positive-class recall at \num{41.18} macro-F1. Three Arabic encoders on the at-risk category did the reverse. They answered positive to almost every call, with positive-class recall of \num{97.4} to \num{100.0} at macro-F1 between \num{33.63} and \num{37.22}. Both kinds of failure therefore land in a similar macro-F1 range of roughly \num{33} to \num{41}. Macro-F1 on its own cannot tell a model that detects no positive case from one that flags every caller. Gaps between ROC-AUC and PR-AUC picked out the same conditions. One Arabic encoder trained for three epochs reached \num{75.13} ROC-AUC with \num{47.24} PR-AUC on high risk, which describes a model that ranks calls acceptably while deciding poorly. Positive-class recall and PR-AUC should therefore be reported alongside macro-averaged measures whenever risk labels are imbalanced.

\begin{table}[!ht]
\caption{Trivial classifiers in both directions give similar macro-F1. Positive-class recall separates them. Macro-F1 and accuracy do not.}\label{tab:collapse}
\begin{tabular}{@{}lllrrr@{}}
\toprule
Condition & category & Direction & Macro-F1 & R+ & Accuracy \\
\midrule
1.5B decoder, Arabic, 3 epochs & High (30) & All negative & 41.18 & 0.00 & 70.00 \\
CAMeLBERT-DA, 3 epochs & At-risk (52) & Near all positive & 37.22 & 100.00 & 53.33 \\
MARBERT-based, 3 epochs & At-risk (52) & All positive & 34.21 & 100.00 & 52.00 \\
MARBERT-based, 10 epochs & At-risk (52) & Near all positive & 33.63 & 97.44 & 50.67 \\
Multilingual BERT, 10 epochs & At-risk (52) & All positive & 34.21 & 100.00 & 52.00 \\
\botrule
\end{tabular}
\end{table}

\section{Discussion}\label{sec:discussion}

Our findings extend prior work on suicide-risk detection in several important ways. Previous studies have largely focused on social media and clinical notes, while the smaller body of helpline research has examined text-based counselling, acoustic features, or structured operator assessments. Our study moves this work to Levantine Arabic crisis calls, using the speech content itself rather than operator-entered fields or acoustic biomarkers. The results are consistent with prior evidence that transformer-based models can identify suicide-related content, but they also show that performance depends strongly on the risk category being detected and the model backbone, rather than following a simple pattern in which more severe risk is necessarily harder to identify. The finding that English translation can preserve classification performance further suggests a potential pathway for applying existing English-language models to Arabic crisis calls, although this requires validation beyond the present dataset. Most importantly, the study extends prior work by demonstrating that these models can be incorporated into a privacy-preserving, within-service pipeline, addressing the need for suicide-risk NLP research that operates in real-world, non-English crisis settings without requiring transfer of raw recordings.

\subsection{
Implications for Suicide-Risk Detection in Arabic Helpline Calls}

Our analysis have shown that sorting de-identified Levantine Arabic helpline transcripts into two risk categories is feasible. For the high-risk category, the models achieved performance sufficient to support further evaluation as a screening aid for helpline operators. Importantly, this performance was achieved within a privacy-preserving pipeline in which the audio remains within the helpline environment. Additionally, we found that lower-severity ideation was more difficult to distinguish than high-risk ideation. This contrasts with the field’s predominant focus on detecting rare, severe outcomes and suggests that distinguishing less severe but clinically meaningful ideation may require greater attention. We have also shown that translating the transcripts into English before classification did not reduce performance. This is notable because the English input undergoes two transformations—automatic speech recognition followed by translation—whereas the Arabic input undergoes only speech recognition. One possible explanation is that the larger volume of English-language pretraining data helps compensate for information lost during these transformations. However, this finding is based on a single sample and translation model and should not be interpreted as evidence that translation will generally improve or preserve performance across settings. In addition, macro-F1 and accuracy alone can mask clinically important failure modes. A model that predicts high risk for nearly every caller and one that predicts no callers as high risk may achieve seemingly moderate aggregate scores while being clinically unusable. Positive-class recall and PR-AUC should therefore be reported alongside macro-averaged metrics to provide a clearer picture of performance on the risk-positive class.

\subsection{Implications for Helpline Practice}

The role of the model is not to determine suicide risk on its own, but to support the people doing this work. For example, the model could flag calls for a supervisor to review, help prioritize callers for follow-up, or remind an operator to complete a structured risk assessment. In this role, detecting positive cases is more important than avoiding false alarms. An unnecessary review may cost a supervisor a few minutes, whereas missing a high-risk caller could have far more serious consequences. Two steps should precede any use in practice. First, performance should be evaluated on consecutive calls as they arrive, rather than on a retrospective sample that may be limited by factors such as call duration. Second, helpline staff should be given estimates of the expected number of additional reviews per week so they can understand the workload implications before the tool is deployed, rather than after implementation.

\subsection{Limitations}\label{subsec:limitations}

All the data come from one helpline in one country over a limited period. We do not know how the model would do in another service, another dialect region or another time. Only \num{545} of the \num{987} de-identified transcripts matched a service record, and a further \num{17} were dropped because duplicate records disagreed. The \num{442} unmatched calls carry no recorded assessment, so we cannot check whether they differ from the calls we kept. Calls longer than ten minutes were outside the eligibility window, so the sample does not cover the longest calls. We did not measure word error rate against transcripts written by hand. Levantine crisis speech is hard to transcribe, with dialect, broken speech, crying and overlapping voices. Both languages inherit those errors, and the English side adds translation error on top. Part of what we are comparing is therefore the quality of the text going in. A small hand-transcribed sample would settle this. The labels are the operator's recorded C-SSRS assessment, not an independent review. A category is positive if any of its items is positive, so one wrong item makes the whole label positive. Errors in the labels therefore push in one direction only. At best the model reproduces operator judgement, including its mistakes. It does not predict suicidal behaviour, which was never observed here.

\subsection{Conclusion}

Sorting privacy-preserved Arabic helpline transcripts into two risk categories is feasible. Performance for the high-risk category supports further evaluation of the model as a potential decision-support tool for helpline operators. Lower-severity ideation was more difficult to distinguish than high-risk ideation. Translating the transcripts into English before classification also produced useful results. Positive-class recall and PR-AUC should be reported alongside macro-averaged metrics because macro-averaged scores alone can obscure poor performance on the positive class.

\backmatter


\section*{Acknowledgements}

\begin{itemize}
\item \textbf{Code availability.} The transcription and de-identification code is available at \url{https://github.com/SarielMa/Arabic_transcribe_deidentify}. The modeling code is available at \url{https://github.com/SarielMa/Arabic_Suicide_project} 
\end{itemize}

\bibliography{references}

\end{document}